\def\year{2022}\relax
\documentclass[letterpaper]{article} 
\usepackage{aaai22}  
\usepackage{times}  
\usepackage{helvet}  
\usepackage{courier}  
\usepackage[hyphens]{url}  
\usepackage{graphicx} 
\usepackage{natbib}  
\usepackage{caption} 
\DeclareCaptionStyle{ruled}{labelfont=normalfont,labelsep=colon,strut=off} 
\usepackage{microtype}
\usepackage{booktabs}
\usepackage{graphicx}
\usepackage{booktabs} 
\usepackage{amsmath,amsthm,amsfonts,amssymb,amscd}
\usepackage{lastpage}
\usepackage{enumerate}
\usepackage{fancyhdr}
\usepackage{mathrsfs}
\usepackage{xcolor}
\usepackage{mathtools}

\usepackage{amssymb}
\usepackage{subcaption}
\usepackage{array}
\usepackage{bbm}
\usepackage{nicefrac}
\usepackage[font={small}]{caption}

\usepackage[ruled,vlined,commentsnumbered,linesnumbered]{algorithm2e}

\usepackage{stackengine} 

\usepackage{amsmath}
\usepackage{amsthm}
\usepackage{bbm}
\DeclareMathOperator*{\argmax}{argmax}

\usepackage{listings}
\usepackage{dsfont}

\usepackage[utf8]{inputenc} 
\usepackage[T1]{fontenc}    
\usepackage{hyperref}       
\usepackage{url}            
\usepackage{booktabs}       
\usepackage{amsfonts}       
\usepackage{nicefrac}       
\usepackage{microtype}      
\usepackage{xcolor}         

\newcommand{\name}{JUMBO}
\newtheorem{theorem}[]{Theorem}
\newtheorem{definition}[]{Definition}

\newtheorem{lemma}[theorem]{Lemma}
\newtheorem{assumption}[]{Assumption}

\DeclareRobustCommand{\rchi}{{\mathpalette\irchi\relax}}

\usepackage{enumitem}

\newcommand{\irchi}[2]{\raisebox{\depth}{$#1\chi$}}

\newcommand{\warmgp}{$\mathcal{GP}^{\text{warm}}$}
\newcommand{\coldgp}{$\mathcal{GP}^{\text{cold}}$}
\newcommand{\offlinedata}{$\mathcal{D}^{\text{aux}}$}
\newcommand{\acqwarm}[1]{$\alpha_{#1}^{\text{warm}}(x)$}
\newcommand{\acqcold}[1]{$\alpha_{#1}^{\text{cold}}(x)$}

\newcommand{\prob}[1]{P\{ #1 \}}

\newcommand{\evAb}{\bar{\mathcal{A}}}
\newcommand{\evBb}{\bar{\mathcal{B}}}

\title{\name{}: Scalable Multi-task Bayesian Optimization using Offline Data}
\author{
    Kourosh Hakhamaneshi\textsuperscript{\rm 1}, Pieter Abbeel \textsuperscript{\rm 1}, Vladimir Stojanovic \textsuperscript{\rm 1}, Aditya Grover\textsuperscript{\rm 2}\\
}
\affiliations{
    \textsuperscript{\rm 1}UC Berkeley, 
    \textsuperscript{\rm 2}FAIR \\

}

\usepackage{bibentry}

\begin{document}

\maketitle

\begin{abstract}
The goal of Multi-task Bayesian Optimization (MBO) is to minimize the number of queries required to accurately optimize a target black-box function, given access to offline evaluations of other auxiliary functions. When offline datasets are large, 
the scalability of prior approaches comes at the expense of expressivity and inference quality.
We propose \name{}, an MBO algorithm that sidesteps these limitations by querying additional data based on a combination of acquisition signals derived from training two Gaussian Processes (GP): a \textit{cold-GP} operating directly in the input domain and a \textit{warm-GP} that operates in the feature space of a deep neural network pretrained using the offline data.
Such a decomposition can dynamically control the reliability of information derived from the online and offline data and the use of pretrained neural networks permits scalability to large offline datasets.
Theoretically, we derive regret bounds for \name{} and show that it achieves no-regret under conditions analogous to GP-UCB~\cite{srinivas2010gaussian}. Empirically, we demonstrate significant performance improvements over existing approaches on two real-world optimization problems: hyper-parameter optimization and automated circuit design.
\end{abstract}

\section{Introduction}
\label{sec:intro}

Many domains in science and engineering involve the optimization of an unknown black-box function. 
Such functions can be expensive to evaluate, due to costs such as time and money.
Bayesian optimization (BO) is a popular framework for such problems as it seeks to minimize the number of function evaluations required for optimizing a target black-box function \cite{shahriari2015taking, frazier2018tutorial}.
In real-world scenarios however, we often have access to offline evaluations of one or more auxiliary black-box functions related to the target function.
For example, one might be interested in finding the optimal hyperparameters of a machine learning model for a given problem and may have access to an offline dataset from previous runs of training the same model on a different dataset for various configurations.
Multi-task Bayesian optimization (MBO) is an optimization paradigm that extends BO to exploit such additional sources of information from related black-box functions for efficient optimization \cite{swersky2013multi}.

Early works in MBO employ multi-task Gaussian Processes (GP) with inter-task kernels to capture the correlations between the auxiliary and target function~\cite{swersky2013multi, williams2007multi, poloczek2016warm}. 
Multi-task GPs however fail to scale to large offline datasets. More recent works have proposed combining neural networks (NN) with probabilistic models to improve scalability.
For example, MT-BOHAMIANN~\cite{springenberg2016bayesian} uses Bayesian NNs (BNN) ~\cite{neal2012bayesian} as surrogate models for MBO.
The performance however, depends on the quality of the inference procedure.
In contrast, MT-ABLR~\cite{perrone2018scalable} uses a deterministic NN followed by a Bayesian Linear Regression (BLR) layer at the output to achieve scalability while permitting exact inference.
However, the use of a linear kernel can limit the expressiveness of the posterior.


We propose \name{}, an MBO algorithm that sidesteps the limitations in expressivity and tractability of prior approaches.
In \name{}, we first train a NN  on the auxiliary data to learn a feature space, akin to MT-ABLR but without the BLR restriction on the output layer.
Thereafter, we train two GPs simultaneously for online data acquisition via BO: a \textit{warm-GP} on the feature space learned by the NN and a \textit{cold-GP} on the raw input space.
The acquisition function in \name{} combines the individual acquisition functions of both the GPs.
It uses the warm-GP to reduce the search space by filtering out poor points.
The remaining candidates are scored by the acquisition function for the cold-GP to account for imperfections in learning the feature space of the warm-GP.
The use of GPs in the entire framework ensures tractability in posterior inference and updates.

Theoretically, we show that \name{} is a no-regret algorithm under conditions analogous to those used for analyzing GP-UCB~\cite{srinivas2010gaussian}. In practice, we observe 
significant 
improvements 
over the closest baseline on two real-world applications: transferring prior knowledge in hyper-parameter optimization 
and automated circuit design. 




\section{Background}
We are interested in maximizing a target black-box function $f: \rchi \to \mathbb{R}$ defined over a discrete or compact set $\rchi \subseteq \mathbb{R}^d$.
We assume only query access to $f$. 
For every query point $x$, we receive a noisy observation $y = f(x) + \epsilon$. Here, we assume $\epsilon$ is standard Gaussian noise, i.e., $\epsilon \sim \mathcal{N}(0, \sigma_{n}^2)$ where $\sigma_n$ is the noise standard deviation.
Our strategy for optimizing $f$ will be to learn a probabilistic model for regressing the inputs $x$ to $y$ using the available data and using that model to guide the acquisition of additional data for updating the model.
In particular, we will be interested in using Gaussian Process regression models within a Bayesian Optimization framework, as described below. 

\subsection{Gaussian Process (GP) Regression}
A Gaussian Process (GP) is defined as a set of random variables such that any finite subset of them follows a multivariate normal distribution.
A GP can be used to define a prior distribution over the unknown function $f$, which can be converted to a posterior distribution once we observe additional data.
Formally, a GP prior is defined by a mean function $\mu_0: \rchi \to \mathbb{R}$ and a valid kernel function $\kappa: \rchi \times \rchi \to \mathbb{R}$.
A kernel function $\kappa$ is valid if it is symmetric and the Gram matrix $K$ is positive semi-definite.
Intuitively, the entries of the kernel matrix  $K_{i,j} = \kappa(x_{i}, x_{j})$ measure the similarity between any two points $x_{i}$ and $x_{j}$.
Given points $X = \{x_1, x_2, \dots, x_n\}$, the distribution of the function evaluations $\mathbf{f} = [f(x_1), f(x_2), \dots f(x_n)]$ in a GP prior follows a normal distribution, such that $\mathbf{f} | X \sim \mathcal{N}(\mu_0(X), K(X, X))$ where $\mu_0(X)=[\mu_0(x_1), \mu_0(x_2), \dots \mu_0(x_n)]$ and $K(X, X)$ is a covariance matrix.
For simplicity, we will henceforth assume $\mu_0$ to be a zero mean function.



Given a training dataset $\mathcal{D}$, let $X_\mathcal{D}$ and $\mathbf{y}_\mathcal{D}$ denote the inputs and their noisy observations. Since the observation model is also assumed to be Gaussian, the posterior over $f$ at a test set of points $X^\ast$ will follow a multivariate normal distribution with the following mean and covariance:

\begin{equation*}
\begin{small}
\begin{aligned}
\mu(\mathbf{f}^*|\mathcal{D}, X^\ast) &= K(X^\ast, X_{\mathcal{D}})^T \Tilde{K}_D^{-1} \mathbf{y}_{\mathcal{D}}, \\
\Sigma(\mathbf{f}^*|\mathcal{D}, X^\ast) &= K(X^*, X^*) - K(X^*, X_{\mathcal{D}})^T\Tilde{K}_D ^{-1} K(X^*, X_{\mathcal{D}}), \\
\text{where } \Tilde{K}_D &= K(X_\mathcal{D},X_\mathcal{D}) + \sigma_n^2I. 
\end{aligned}
\end{small}
\end{equation*}


Due to the inverse operation during posterior computation, standard GPs can be computationally prohibitive for modeling large datasets.
We direct the reader to \cite{rasmussen2003gaussian} for an overview on GPs.

\subsection{Bayesian Optimization (BO)}
Bayesian Optimization (BO) is a class of sequential algorithms for sample-efficient optimization of expensive black-box functions \cite{frazier2018tutorial,shahriari2015taking}.
A BO algorithm typically runs for a fixed number of rounds.
At every round $t$, the algorithm selects a query point $x_t$ and observes a noisy function value $y_t$.
To select $x_t$, the algorithm first infers the posterior distribution over functions $p(\mathbf{f} \vert \{(x_i, y_i)\}_{i=1}^{t-1})$ via a probabilistic model (e.g., Gaussian Processes). Thereafter, $x_t$ is chosen to optimize an uncertainty-aware acquisition function that balances exploration and exploitation. 
For example, a popular acquisition function is the Upper Confidence Bound (UCB) which prefers points that have high expected value (exploitation) and high uncertainty (exploration).
With the new point $(x_t, y_t)$, the posterior distribution can be updated and the whole process is repeated in the next round.

At round $t$, we define the instantaneous regret as $r_t = f(x^*) - f(x_t)$ where $x^\ast$ is the global optima and $x_t$ maximizes the acquisition function. Similarly, we can define the cumulative regret at round $T$ as the sum of instantaneous regrets $R_T= \sum_{t=1}^T r_{t}$. A desired property of any BO algorithms is to be \textit{no-regret} where the cumulative regret is sub-linear in $T$ as $T\to \infty$, i.e., $\lim_{T\to\infty} \nicefrac{R_T}{T} = 0$.

\subsection{Multi-task Bayesian Optimization (MBO)}
\label{sec:background_mtbo}
Our focus setting in this work is a variant of BO, called Multi-Task Bayesian Optimization (MBO). Here, we assume $K>0$ auxiliary real-valued black-box functions $\{f_1, \dots, f_K\}$, each having the same domain $\rchi$ as the target function $f$ \cite{swersky2013multi, springenberg2016bayesian}.
For each function $f_k$, we have an offline dataset $\mathcal{D}^{(k)}$ consisting of pairs of input points $x$ and the corresponding function evaluations $f_k(x)$. If these auxiliary functions are related to the target function, then we can transfer knowledge from the offline data $\mathcal{D}^{\text{aux}} = \mathcal{D}^{(1)} \cup \dots \cup \mathcal{D}^{(K)}$ to improve the sample-efficiency for optimizing $f$. 
In certain applications, we might also have access to offline data from $f$ itself. However, in practice, $f$ is typically expensive to query and its offline dataset $\mathcal{D}^f$ will be very small.

We discuss some prominent works in MBO that are most closely related to our proposed approach below. 
See Section~\ref{sec:related} for further discussion about other relevant work.

\textbf{Multi-task BO}~\cite{swersky2013multi} is an early approach that employs a custom kernel within a multi-task GP~\cite{williams2007multi} to model the relationship between the auxiliary and target functions.
Similar to standard GPs, multi-task GPs fail to scale for large offline datasets.

On the other hand, parametric models such as neural networks (NN), can effectively scale to larger datasets but do not defacto quantify uncertainty. 
Hybrid methods such as \textbf{DNGO}~\cite{snoek2015scalable} achieve scalability for (single task) BO through the use of a feed forward deep NN followed by Bayesian Linear Regression (BLR) \cite{bishop2006pattern}.
The NN is trained on the existing data via a simple regression loss (e.g, mean squared error).
Once trained, the NN parameters are frozen and the output layer is replaced by BLR for the BO routine.
For BLR, the computational complexity of posterior updates scales linearly with the size of the dataset.
This step can be understood as applying a GP to the output features of the NN with a linear kernel (i.e. $\kappa(x_i, x_j) = h_\phi(x_i)^T h_\phi(x_j)$ where $h$ is the NN function with parameters $\phi$).
For BLR, the computational complexity of posterior inference is linear w.r.t. the number of data points and thus DNGO can scale to large offline datasets.

\textbf{MT-ABLR}~\cite{perrone2018scalable} extends DNGO to multi task settings by training a single NN to learn a shared representation $h_\phi(x)$ followed by task-specific BLR layers (i.e. predicting $f_1(x), ..., f_K(x)$, and $f(x)$ based on inputs).
The learning objective corresponds to the maximization of sum of the marginal log-likelihoods for each task: $\sum_{t=1}^{K + 1} p(\mathbf{y}_t | w_t, h_\phi(X_t), \sigma_t)$.
The main task is included in the last index, $w_t$ is the Bayesian Linear layer weights for task $t$ with prior $p(w_t) = \mathcal{N}(0, \sigma_{w_t}^2I)$, $\sigma_t$ and $\sigma_{w_t}$ are the hyper-prior parameters, and $(X_t, \mathbf{y}_t)$ is the observed data from task $t$. Learning $h_\phi(x)$ by directly maximizing the marginal likelihood improves the performance of DNGO while maintaining the computational scalability of its posterior inference in case of large offline data. However, both DNGO and ABLR have implicit assumptions on the existence of a feature space under which the target function can be expressed as a linear combination. This can be a restrictive assumption and furthermore, there is no guarantee that given finite data such feature space can be learned.



\textbf{MT-BOHAMIANN}~\cite{springenberg2016bayesian} addresses the limited expressivity of prior approaches by employing Bayesian NNs to specify the posterior over $\mathbf{f}$ and feed the NN with input $x$ and additional learned task-specific embeddings $\psi(t)$ for task $t$. 
While allowing for a principled treatment of uncertainties, fully Bayesian NNs are computationally expensive to train and their performance depends on the approximation quality of stochastic gradient HMC methods used for posterior inference.


\section{Scalable MBO via \name{}}
\label{sec: our_approach}
In the previous section, we observed that prior MBO algorithms make trade-offs in either posterior expressivity or inference quality in order to scale to large offline datasets.
Our goal is to show that these trade-offs can be significantly mitigated and consequently, the design of our proposed MBO framework, which we refer to as \textbf{J}oint \textbf{U}pper confidence \textbf{M}ulti-task \textbf{B}ayesian \textbf{O}ptimization (\name{}), will be guided by the following desiderata: (1) Scalability to large offline datasets (e.g., via NNs) (2) Exact and computationally tractable posterior updates (e.g., via GPs) (3) Flexible and expressive posteriors (e.g., via non-linear kernels).

\subsection{Regression Model}
\label{sec:regression}

\begin{figure}
    \centering
    \includegraphics[width=\linewidth]{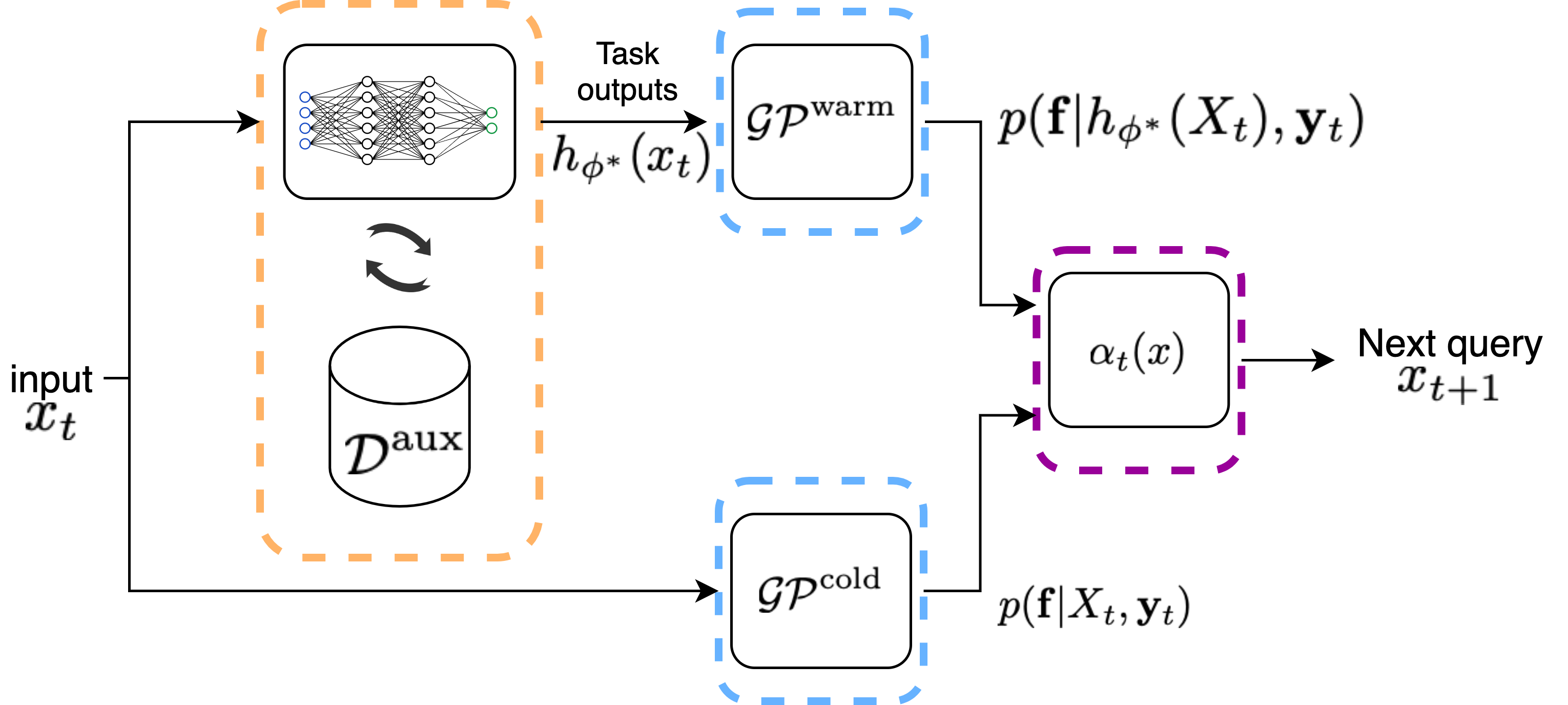}
    \caption{
    \name{}. During the pretraining phase, we learn a NN mapping $h_{\phi^\ast}$ (orange) for the warm-GP. The next query based on $\alpha_t$ (purple) will be the point that has a high score based on the acquisition function of both warm and cold GP (blue). 
    }
    \label{fig:flow}
\end{figure}

The regression model in \name{} is composed of two GPs: a \textit{warm-GP} and a \textit{cold-GP} denoted by \warmgp$(0, \kappa^w)$ and \coldgp$(0, \kappa^c)$, respectively.
As shown in Figure~\ref{fig:flow}, both GPs are trained to model the target function $f$ but operate in different input spaces, as we describe next.

\warmgp{} (with hyperparameters $\theta_w$) operates on a feature representation of the input space $h_\phi(x)$ derived from the offline dataset \offlinedata.
To learn this feature space, we train a multi-headed feed-forward NN to minimize the mean squared error for each auxiliary task, akin to DNGO~\cite{snoek2015scalable}. 
Thereafter, in contrast to both DNGO and ABLR, we do not train separate output BLR modules.
Rather, we will directly train \warmgp{} on the output of the NN using the data acquired from the target function $f$.
Note that for training \warmgp, we can use any non-linear kernel, which results in an expressive posterior that allows for exact and tractable inference using closed-form expressions.

Additionally, we can encounter scenarios where some of the auxiliary functions are insufficient in reducing the uncertainty in inferring the target function.
In such scenarios, relying solely on \warmgp{}  can significantly hurt performance. 
Therefore, we additionally initialize \coldgp{}  (with hyperparameters $\theta_c$) directly on the input space $\rchi$.

If we also have access to offline data from $f$ (i.e. $\mathcal{D}^f$), the hyperparameters of the warm and cold GPs can also be pre-trained jointly along with the neural network parameters. The overall pre-training objective is then given by:

\begin{equation}
    \label{eq:loss}
    \resizebox{0.9\hsize}{!}{
    $\mathcal{L}(\phi, \theta_w, \theta_c) = \mathcal{L}^{\text{MSE}}(\phi \vert \mathcal{D}^{\text{aux}}) + \mathcal{L}^{\mathcal{GP}}(\theta_w \vert \mathcal{D}^f) + \mathcal{L}^{\mathcal{GP}}(\theta_c \vert \mathcal{D}^f)$
    }
\end{equation}


where $\mathcal{L}^{\mathcal{GP}}( \cdot \vert \mathcal{D}^f)$ denotes the negative marginal log-likelihood for the corresponding GP on $\mathcal{D}^f$.

\subsection{Acquisition Procedure}
Post the offline pre-training of the \name{}'s regression model, we can use it for online data acquisition in a standard BO loop.
The key design choice here is the acquisition function, which we describe next.
At round $t$, let \acqwarm{t} and \acqcold{t} be the single task acquisition function (e.g. UCB) of the warm and cold GPs, after observing $t-1$ data points, respectively.

Our guiding intuition for the acquisition function in \name{} is that we are most interested in querying points which are scored highly by both acquisition functions. Ideally, we want to first sort points based on $\alpha^{\text{warm}}$ scores and then from the top choices select the ones with highest $\alpha^{\text{cold}}$ score. 
To realize this acquisition function on a continuous input domain, we define it as a convex combination of the individual acquisition functions by employing a \textit{dynamic} interpolation coefficient $\lambda_{t}(x) \in [0, 1]$. Formally, 

\begin{align}
    \label{eq:acq}
    \alpha_{t}(x) = \lambda_t(x) \alpha^{\text{cold}}_{t}(x) + (1 - \lambda_t(x)) \alpha^{\text{warm}}_{t}(x).
\end{align}

In Eq.~\ref{eq:acq}, 
By choosing $\lambda_t(x)$ to be close to $1$ for points with $\alpha_t^{\text{warm}}(x) \approx \max_x \alpha_t^{\text{warm}}(x)$, we can ensure to acquire points that have high acquisition scores as per both \acqcold{t} and \acqwarm{t}. Next, we will discuss some theoretical results that shed more light on the design of $\lambda_t(x)$.

\subsection{Theoretical Analysis}
\label{sec:theory}
Here, we will formally derive the regret bound for \name{} and provide insights on the conditions under which \name{} outperforms GP-UCB~\cite{srinivas2010gaussian}.
For this analysis, we will use Upper Confidence Bound (UCB) as our acquisition function for the warm and cold GPs. To do so, we utilize the notion of Maximum Information Gain (MIG).

\begin{definition}[Maximum Information Gain \cite{srinivas2010gaussian}]
Let $f \sim \mathcal{GP}(0, \kappa)$, $\kappa: \mathbb{R}^d \times \mathbb{R}^d \to \mathbb{R}$. Consider any $\rchi \subset \mathbb{R}^d$ and let $\Tilde{\rchi} = \{x_1, ..., x_n\} \subset \rchi$ be a finite subset. Let $\mathbf{y}_{\Tilde{\rchi}} \in \mathbb{R}^n$ be $n$ noisy observations such that $(\mathbf{y}_{\Tilde{\rchi}})_i = (\mathbf{f}_{\Tilde{\rchi}})_i + \epsilon_i$, $\epsilon_i \sim \mathcal{N}(0, \sigma_n^2)$. Let $I$ denote the Shannon mutual information. 

The MIG $\Psi_{n}(\rchi)$ of set $\rchi$ after $n$ evaluations is the maximum mutual information between the function values and observations among all choices of n points in $\rchi$. Formally, 

\begin{equation*}
    \Psi_{n}(\rchi) = \max_{\Tilde{\rchi} \subset \rchi, |\Tilde{\rchi}| = n} I(\mathbf{y}_{\Tilde{\rchi}}; \mathbf{f}_{\Tilde{\rchi}})
\end{equation*}

\end{definition}

This quantity depends on kernel parameters and the set $\rchi$, and also serves as an important tool for characterizing the difficulty of a GP-bandit. For a given kernel, it can be shown that $\Psi_{n}(\rchi) \propto \Pi(\rchi)$ where $\Pi(\rchi) = |\rchi|$ for discrete and $\text{Vol}(\rchi)$ for the continuous case~\cite{srinivas2010gaussian}. 
For example for \textit{Radial Basis} kernel $\Psi_n([0, 1]^d) \in O(\log(n)^{d+1})$.
For brevity, we focus on settings where $\rchi$ is discrete. 
Further results and analysis for the continuous case are deferred to Appendix~\ref{app:proofs}.

For GP-UCB \cite{srinivas2010gaussian}, 
it has been shown that for any $\delta \in (0, 1)$,
if $f \sim \mathcal{GP}(0, \kappa)$ (i.e., the GP assigns non-zero probability to the target function $f$), then the cumulative regret $R_T$ after $T$ rounds will be bounded with probability at least $1-\delta$:
\begin{equation}
    Pr\{R_T \le \sqrt{CT\beta_T\Psi_T(\rchi)}, \forall T \ge 1\} \ge 1 - \delta
\end{equation}

with $C = \frac{8}{\log(1 + \sigma_n ^ {-2})}$ and $\beta_T = 2\log\left(\frac{|\rchi| \pi^2 T^2}{6\delta}\right)$. 

Recall that $h_\phi: \rchi \to \mathcal{Z}$ is a mapping from input space $\rchi$ to the feature space $\mathcal{Z}$.
We will further make the following modeling assumptions to ensure that the target black-box function $f$ is a sample from both the cold and warm GPs.

\begin{assumption}
    \label{ass:coldgp}
    $f \sim \mathcal{GP}^{\text{cold}}(0, \kappa^{c})$.
\end{assumption}

\begin{assumption}
    \label{ass:warmgp}
    Let $\phi^\ast$ denote the NN parameters obtained via pretraining (Eq.~\ref{eq:loss}).Then, there exists a function $g \sim \mathcal{GP}^{\text{warm}}(0, \kappa^{w})$ such that $f = g \circ h_{\phi^\ast}$.
\end{assumption}

\begin{theorem}
\label{theorem:1}
Let $\rchi_g \subset \rchi$ and $\bar{\rchi}_g = \rchi \setminus \rchi_g$ be some arbitrary partitioning of the input domain $\rchi$. Define the interpolation coefficient as an indicator $\lambda_t(x) = \mathbbm{1}(x\in\rchi_g)$.
Then under Assumptions \ref{ass:coldgp} and \ref{ass:warmgp}, \name{} is no-regret. 

Specifically, let $s$ be the number of rounds such that the \name{} queries for points $x_t \in \bar{\rchi}_g$.
Then, for any  $\delta \in (0,1)$, running \name{} for $T$ iterations results in a sequence of candidates $(x_t)_{t=1}^{t=T}$ for which the following holds with probability at least $1 - \delta$:

\begin{equation}
    \label{eq:regret}
    R_{T} < \sqrt{CT\beta_T \{\Psi_{T-s}(\rchi_g) + \Psi_{s}(\bar{\mathcal{Z}}_g)\}}, \forall T \ge 1
\end{equation}

where $C = \frac{8}{\log\left(1 + \sigma_{n}^{-2}\right)}$,
$\beta_t = 2 \log\left(\frac{|\rchi| \pi ^ 2 t ^2}{3\delta}\right)$, 
and 
$\bar{\mathcal{Z}}_g = \{h_{\phi^\ast}(x) | x \in \bar{\rchi}_g\}$ is the set of output features for $\bar{\rchi}_g$.
\end{theorem}

\begin{figure}
    \centering
    \includegraphics[width=0.7\linewidth]{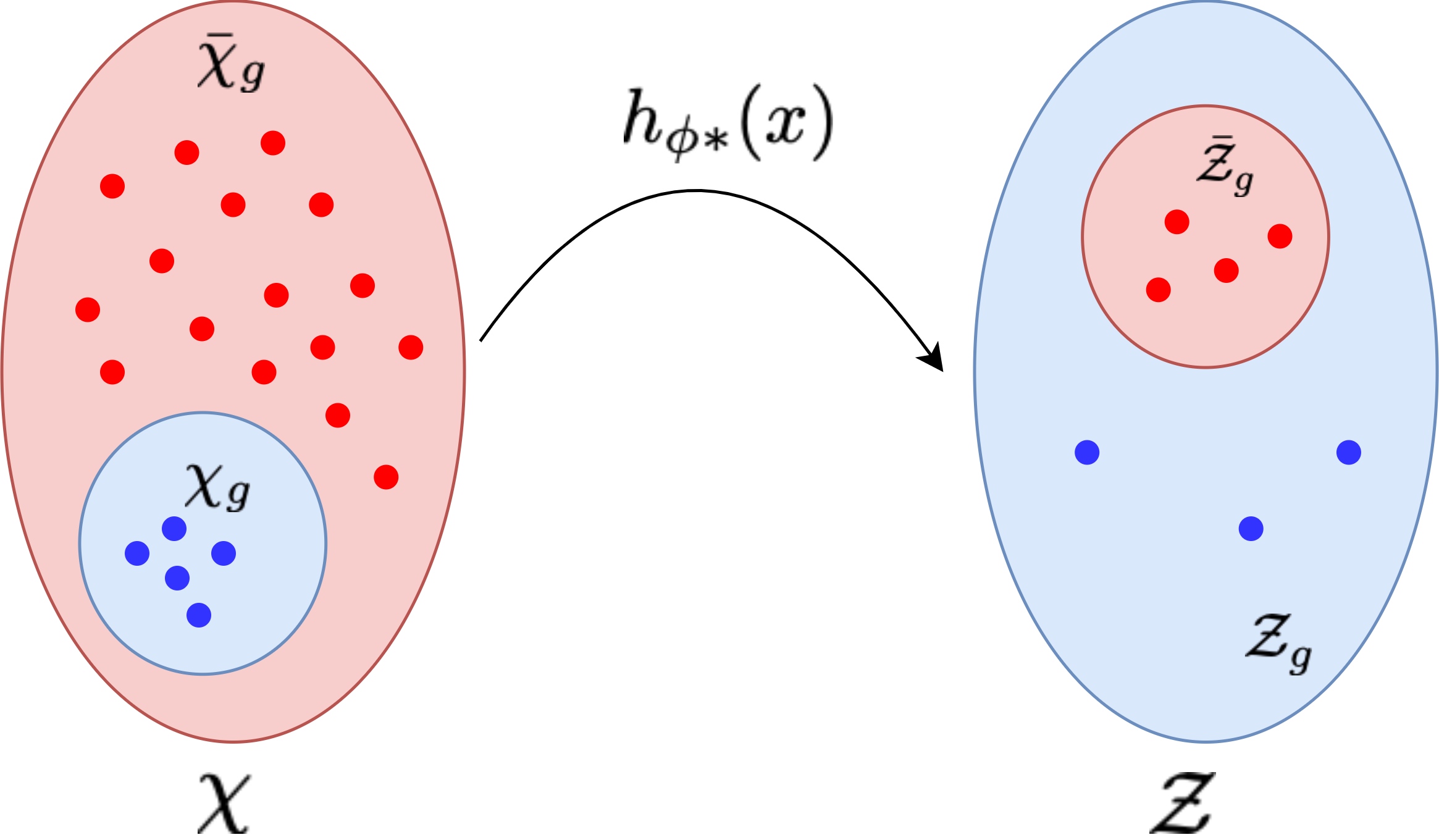}
    \caption{The effect of the pre-trained NN $h_{\phi^\ast}(x)$ on $\rchi$. In the desirable case, $\bar{\rchi}_g$ gets significantly compressed  to $\bar{\mathcal{Z}}_g$.
    }
    \label{fig:venn}
\end{figure}

Based on the regret bound in Eq.~\ref{eq:regret}, we can conclude that if the partitioning $\rchi_g$ is chosen such that $\Pi(\bar{\mathcal{Z}}_g) \ll \Pi(\bar{\rchi}_g)$ \textit{and} $\Pi(\rchi_g) \ll \Pi(\rchi)$, then \name{} has a tighter bound than GP-UCB. The first condition implies that the second term in Eq.~\ref{eq:regret} is negligible and intuitively means that \warmgp{} will only need a few samples to infer the posterior of $f$ defined on $\bar{\rchi}_g$, making BO more sample efficient. The second condition implies that the $\Psi_{T-s}(\rchi_g) \ll \Psi_T(\rchi)$ which in turn makes the regret bound of \name{} tighter than GP-UCB. Note that $\rchi_g$ cannot be made arbitrarily small, since $\bar{\rchi}_g$ (and therefore $\bar{\mathcal{Z}}_g$) will get larger which conflicts with the first condition.

Figure \ref{fig:venn} provides an illustrative example. If the learned feature space $h_{\phi^\ast}(x)$ compresses set $\bar{\rchi}_g$ to a smaller set $\bar{\mathcal{Z}}_g$, then \warmgp{}  can infer the posterior of $g(h_{\phi^\ast}(x))$ with only a few samples in $\bar{\rchi}_g$ (because MIG is lower). Such $h_{\phi^\ast}(x)$ will likely emerge when tasks share high-level features with one another. In the appendix, we have included an empirical analysis to show that \warmgp is indeed operating on a compressed space $\mathcal{Z}$. Consequently, if $\rchi_g$ is reflective of promising regions consisting of near-optimal points i.e. $\rchi_g = \{x\in \rchi \mid f(x^\ast) - f(x) \le l_f\}$ for some $l_f > 0$, BO will be able to quickly discard points from subset $\bar{\rchi}_g$ and acquire most of its points from $\rchi_g$.

\begin{figure*}[htb]
    \centering
    \begin{subfigure}{0.7\textwidth}
        \begin{subfigure}{0.49\textwidth}
            \captionsetup{justification=centering}
            \includegraphics[width=\textwidth]{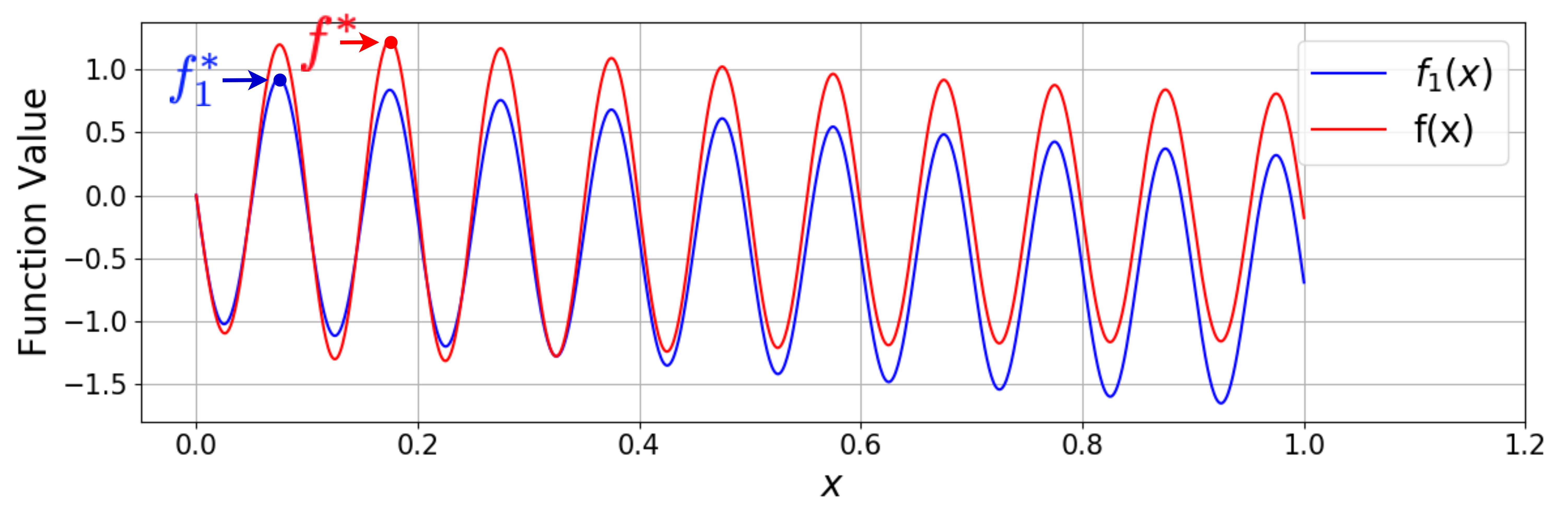}\par\caption{Target (red) and auxiliary (blue) task}\label{fig:ped1}
        \end{subfigure}
        \hfill
        \begin{subfigure}{0.49\textwidth}
            \includegraphics[width=\textwidth]{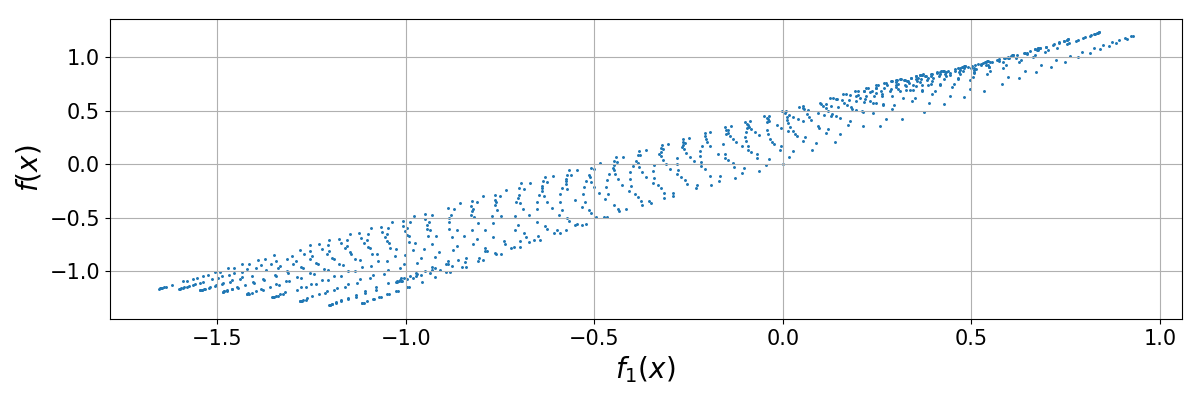}\par\caption{Correlation between target and auxiliary tasks}\label{fig:ped2}
        \end{subfigure}
    \end{subfigure}
    \begin{subfigure}{0.75\textwidth}
        {\captionsetup{justification=centering}
        \includegraphics[width=\textwidth]{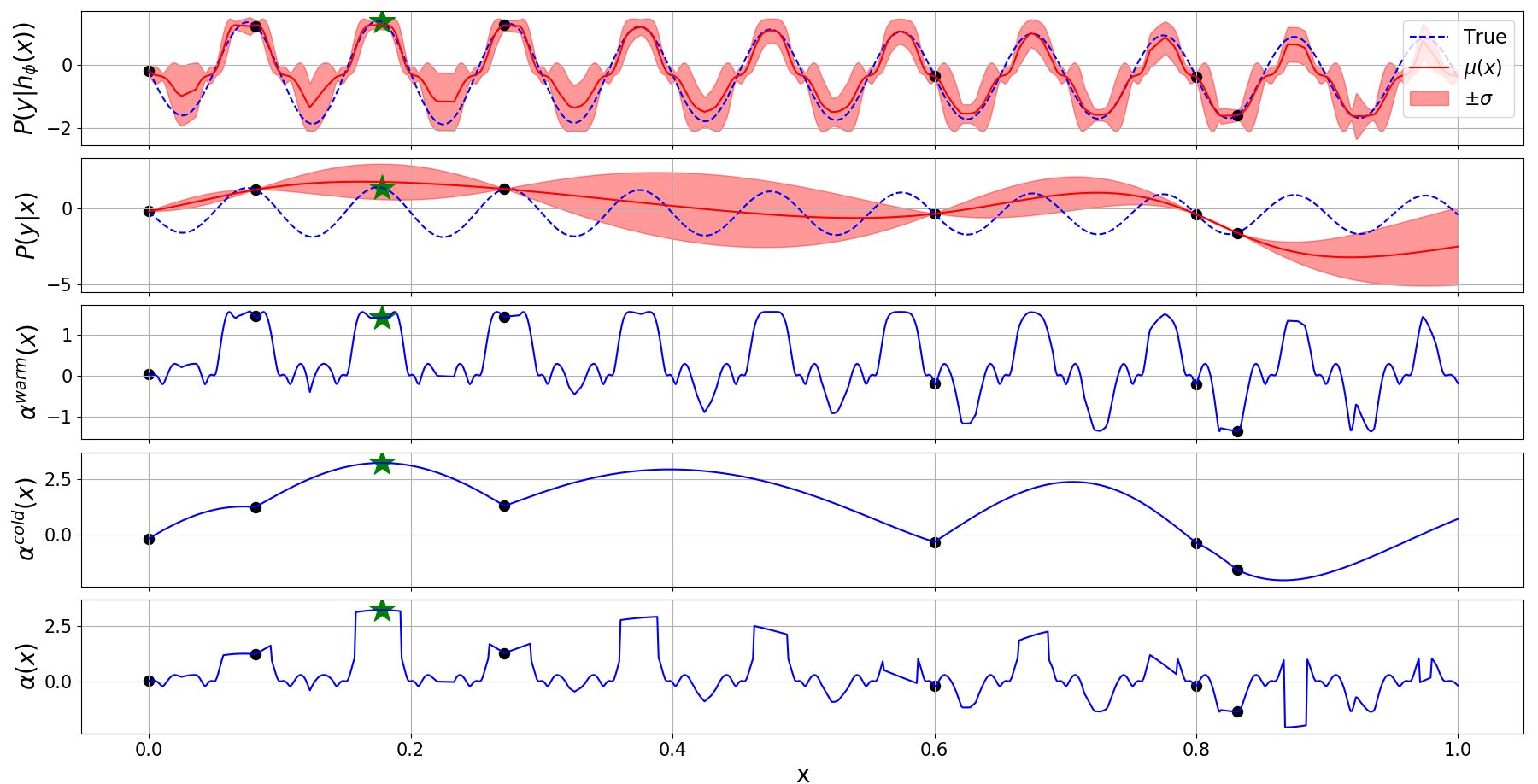}\par\caption{Posterior of \warmgp{}, \coldgp{}, their UCB and \name{}'s acquisition function} \label{fig:ped3}}
    \end{subfigure}
    \caption{
    Dynamics of \name{} after observing 6 data points (a) The two functions have different optimums (b) The tasks are related (c) Iteration 4 of the BO with our proposed model, from top to bottom: (1) GP modeling input to objective using ($x, h_{\phi^\ast}(x), y$) samples (2) GP modeling input to objective using ($x, y$) samples (3) UCB acquisition function for \warmgp (4) UCB acquisition function for \coldgp (5) \name{}'s acquisition function that compromises between the optimum of the two.
    }
    \label{fig:ped}
\end{figure*}

\subsection{Choice of interpolation coefficient $\lambda_t(x)$}
The above discussion suggests that the partitioning $\rchi_g$ should ideally consist of near-optimal points. In practice, we do not know $f$ and hence, we rely on our surrogate model to define $\rchi_g^{(t)} = \{x \in \rchi \mid \alpha_t^{\text{warm}\ast} - \alpha_t^{\text{warm}}(x) \le l_\alpha\}$. 
Here, $\alpha_t^{\text{warm}\ast}$ is the optimal value of \acqwarm{t} and the  acquisition threshold $l_\alpha > 0$ is a hyper-parameter used for defining near-optimal points w.r.t. \acqwarm{t}. At one extreme, $l_\alpha \to \infty$ corresponds to the case where $\alpha_t(x) = \alpha_t^{\text{cold}}(x)$ (i.e. the GP-UCB routine) and the other extreme $l_\alpha \to 0$ corresponds to case with $\alpha_t(x) = \alpha_t^{\text{warm}}(x)$.

Figure~\ref{fig:ped} illustrates a synthetic 1D example on how \name{} obtains the next query point. 
Figure~\ref{fig:ped1} shows the main objective $f(x)$ (red) and the auxiliary task $f_1(x)$ (blue). They share a periodic structure but have different optimums.
Figure \ref{fig:ped2} shows the correlation between the two. 

Applying GP-UCB \cite{srinivas2010gaussian} will require a considerable amount of samples to learn the periodic structure and the optimal solution. However in \name{}, as shown in Figure~\ref{fig:ped3}, the warm-GP, trained on ($h_{\phi^\ast}(x), y$) samples, can learn the periodic structure using only 6 samples, while the posterior of the cold-GP has not yet learned this structure. 

It can also be noted from Figure~\ref{fig:ped3} that  \name{}'s acquisition function is \acqcold{t} when the value of \acqwarm{t} is close to $\alpha_t^{\text{warm}\ast}$. Therefore, the next query point (marked with a star) has a high score based on both acquisition functions. We summarize \name{} in Algorithm~\ref{alg:main}.



\begin{algorithm}[ht]
\caption{\name{}}
\label{alg:main}
\SetAlgoLined

\KwIn{Offline auxiliary dataset $\mathcal{D}^{\text{aux}}$
, Offline target dataset $\mathcal{D}_0^{f}$ (optional; default: empty set), Threshold $l_\alpha$}
\KwOut{Sequence of solution candidates $\{x_t\}_{t=1}^T$ maximizing target function $f$}
\BlankLine

Initialize NN $h_{\phi}(x)$, $\mathcal{GP}^{\text{cold}}$, $\mathcal{GP}^{\text{warm}}$.

Pretrain NN params jointly with $\mathcal{GP}^{\text{cold}}$ and $\mathcal{GP}^{\text{warm}}$ hyper-params using $\mathcal{D}^{\text{aux}}$ and $\mathcal{D}_0^{f}$ as per Eq.~\ref{eq:loss}.\;

Initialize $\mathcal{D}_0^{\text{cold}} =\{\}$, $\mathcal{D}_0^{\text{warm}} = \{\}$.

\For{{round $t=1$ \KwTo $T$}}{

Set $\alpha_{t}^{\text{warm}\ast} = \argmax_{x \in \rchi} \alpha_{t}^{\text{warm}}(x)$.

Set $\lambda_{t}(x) = \mathbbm{1}(\alpha_{t}^{\text{warm}\ast} - \alpha_{t}^{\text{warm}}(x) \le l_\alpha)$.

Set $\alpha_{t}(x) = \lambda_{t}(x)\alpha^{\text{cold}}_{t}(x) + (1-\lambda_{t}(x))\alpha^{\text{warm}}_{t}(x)$

Pick $x_t = \argmax_{x \in \rchi} \alpha_{t}(x)$.

Obtain noisy observation $y_t$ for $x_t$.

Update $\mathcal{D}^{\text{cold}}_t \leftarrow \mathcal{D}^{\text{cold}}_{t-1} \cup \{(x_t, y_t)\}$ and $\mathcal{GP}^{\text{cold}}$.

Update $\mathcal{D}^{\text{warm}}_t \leftarrow \mathcal{D}^{\text{warm}}_{t-1} \cup \{(h_{\phi^\ast}(x_i), y_i)\}$ and $\mathcal{GP}^{\text{warm}}$.

}
\end{algorithm}


\section{Experiments}
\label{sec:experiment}
We are interested in investigating the following questions: (1) How does \name{} perform on benchmark real-world black-box optimization problems relative to baselines? (2) How does the choice of threshold $l_\alpha$ impact the performance of \name{}? (3) Is it necessary to have a non-linear mapping on the features learned from the offline dataset or a BLR layer is sufficient?

Our codebase is based on BoTorch \cite{balandat2020botorch}
and is provided in the Supplementary Materials with additional details in Appendix~\ref{app:exp}.



\subsection{Application: Hyperparameter optimization} 
\label{sec:exp_hpo}
\noindent\textbf{Datasets.}
We consider the task of optimizing hyperparameters for fully-connected NN architectures on 4 regression benchmarks from HPOBench~\cite{klein2019tabular}: \textit{Protein Structure}~\cite{rana2013physicochemical}, \textit{Parkinsons Telemonitoring}~\cite{tsanas2010enhanced}, \textit{Naval Propulsion}~\cite{coraddu2016machine}, and \textit{Slice Localization}~\cite{graf20112d}.
HPOBench provides a look-up-table-based API for querying the validation error of all possible hyper-parameter configurations for a given regression task. 
These configurations are specified via 9 hyperparameters, that include 
continuous, categorical, and integer valued variables.



The objective we wish to minimize is the validation error of a regression task after 100 epochs of training. 
For this purpose, we consider an offline dataset that consists of validation errors for some randomly chosen configurations after 3 epochs on a given dataset. The target task is to optimize this error after 100 epochs.
In \cite{klein2019tabular}, the authors show that this problem is non-trivial as there is small correlation between epochs 3 and 100 for top-1\% configurations across all datasets of interest.

\begin{figure*}[htb]
    \centering
    \includegraphics[width=0.75\linewidth]{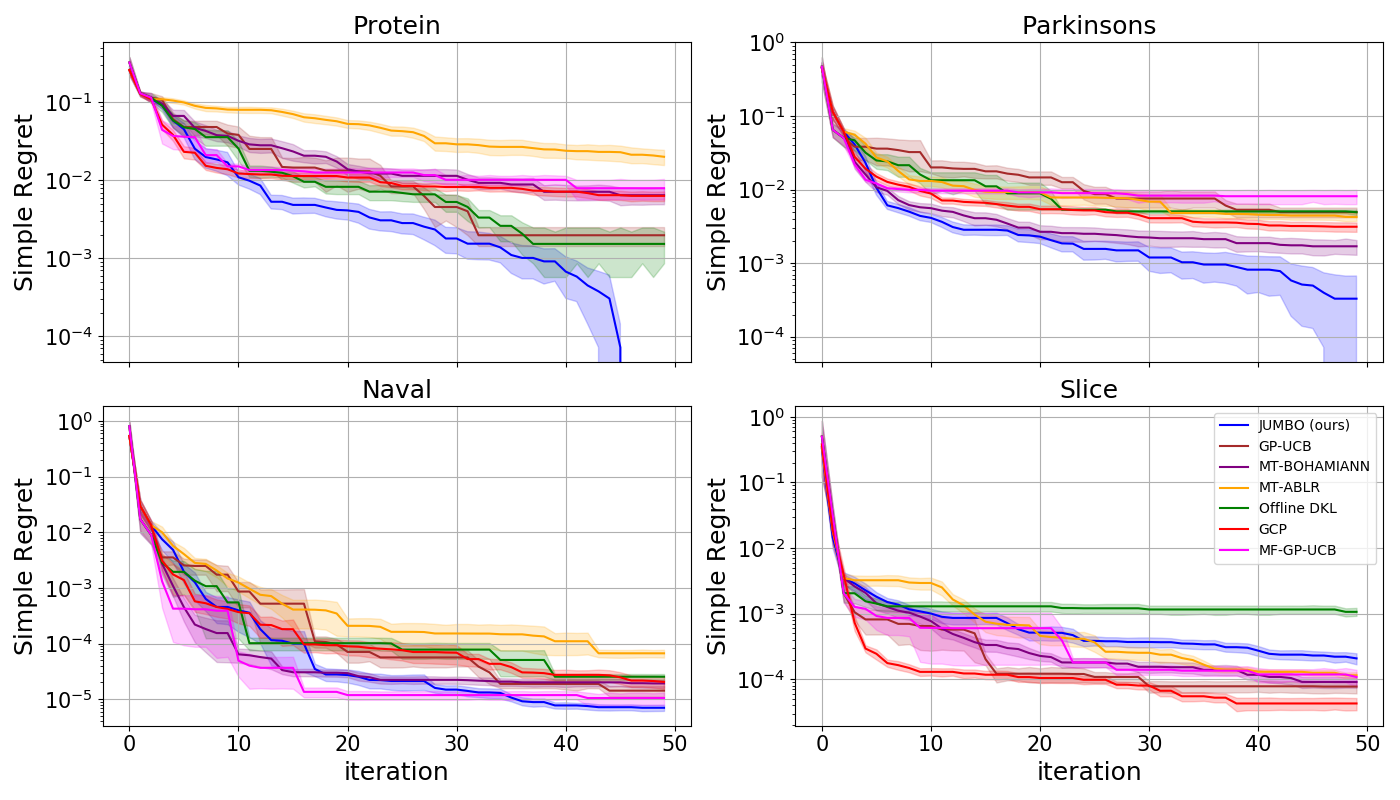}
    \caption{The regret of MBO algorithms 
    on Protein, Parkinsons, Naval, and Slice datasets.
    Standard errors are measured across 20 random seeds.}
    \label{fig:res_main}
\end{figure*}

\noindent\textbf{Evaluation protocol.}
We validate the performance of \name{} against the following baselines with a UCB acquisition function \cite{srinivas2010gaussian}:
\begin{itemize}[leftmargin=*]
\itemsep0em
\item \small\textbf{GP-UCB}~\cite{srinivas2010gaussian} (i.e. cold-GP only) trains a GP from scratch disregarding \offlinedata{} completely. Equivalently, it can be interpreted as \name{} with $\lambda_t(x)=1 \;\;\forall x,t\ge1$ in Eq.~\ref{eq:acq} and $\alpha(x) = \alpha^{\text{UCB}}(x)$.
\item \textbf{MT-BOHAMIANN}~\cite{springenberg2016bayesian} trains a BNN on all tasks jointly via SGHMC (Section~\ref{sec:background_mtbo}).
\item \textbf{MT-ABLR}~\cite{perrone2018scalable} trains a shared NN followed by task-specific BLR layers (Section~\ref{sec:background_mtbo}). 
\item \textbf{GCP} \cite{salinas2020quantile} uses Gaussian Copula Processes to jointly model the offline and online data. 

\item \textbf{MF-GP-UCB} \cite{kandasamy2019multi} extends the GP-UCB baseline to a multi-fidelity setting where the source task can be interpreted as a low-fidelity proxy for the target task.

\item \textbf{Offline DKL} (i.e. warm-GP only) 
is our proposed extension to Deep Kernel Learning, where we train a single GP online in the latent space of a NN pretrained on \offlinedata{} (See Section~\ref{sec:related} for details). 
Equivalently, it can be interpreted as \name{} with $\lambda_t(x)=0 
$ in Eq.~\ref{eq:acq}.

\end{itemize}

\noindent \textbf{Results.}
We run \name{} (with $l_\alpha=0.1$) on all baselines for 50 rounds and 5 random seeds each and measure the simple regret per iteration.
The regret curves are shown in Figure~\ref{fig:res_main}.
We find that \name{} achieves lower regret than the previous state-of-the-art algorithms for MBO in almost all cases.
We believe the slightly worse performance on the slice dataset relative to other baselines is due to the extremely low top-1\% correlation between epoch 3 and epoch 100 on this dataset as compared to others (See Figure 10 in \cite{klein2019tabular}), which could result in a suboptimal search space partitioning obtained via the warm-GP.
For all other datasets, we find \name{} to be the best performing method.
Notably, on the Protein dataset, \name{} is always able to find the global optimum, unlike the other approaches. 


\begin{figure*}[tbh]
    \centering
    \begin{subfigure}{\textwidth}
        \begin{subfigure}{0.32\textwidth}
            \includegraphics[width=\textwidth]{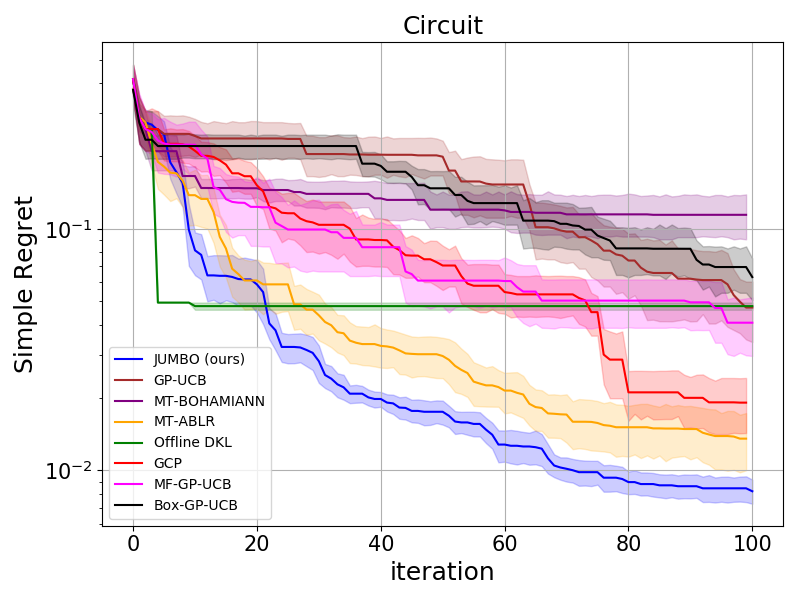}\par
            \caption{}\label{fig:res_ckt}
        \end{subfigure}
        \hfill
        \begin{subfigure}{0.32\textwidth}
            \includegraphics[width=\textwidth]{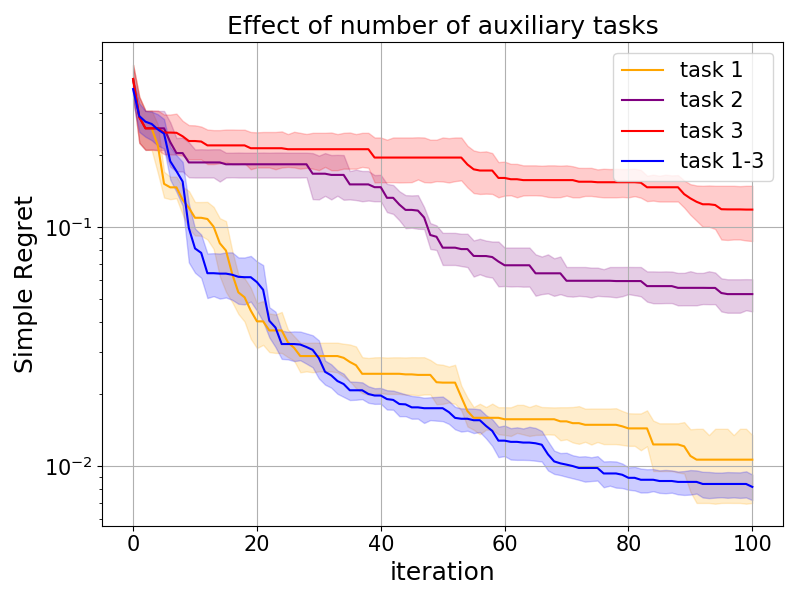}\par
            \caption{}\label{fig:abl_ntask}
        \end{subfigure}
        \hfill
        \begin{subfigure}{0.32\textwidth}
            \includegraphics[width=\textwidth]{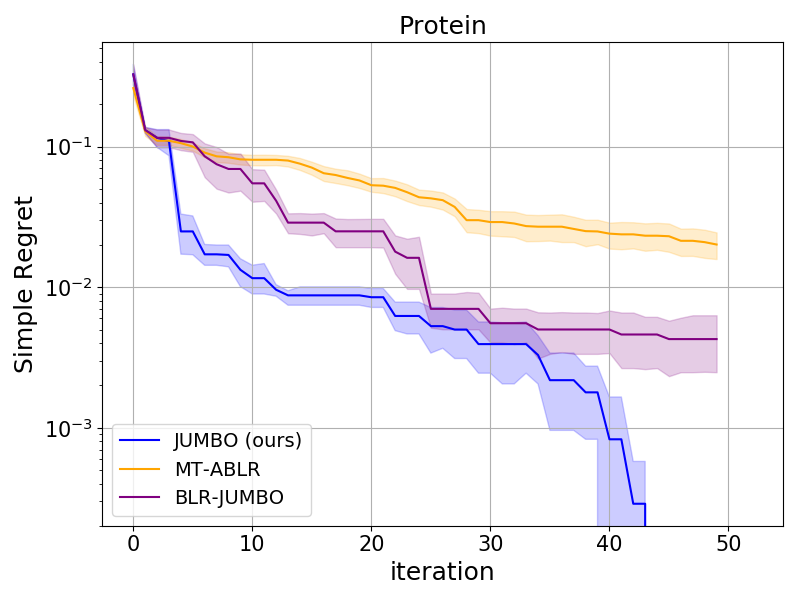}\par
            \caption{}\label{fig:ablation-blr}
        \end{subfigure}
        \hfill    
    \end{subfigure}
    \caption{
    (a) Circuit Design results
    (b) \name{} with 3 aux. tasks is better than \name{} with each individual aux. tasks.
    (c) The Non-linear mapping is a crucial piece of \name{}'s algorithm.
    }
\end{figure*}

\subsection{Application: Automated Circuit Design}
\label{sec:exp_ckt}
Next, we consider a real-world use case in optimizing circuit design configurations for a suitable performance metric, e.g., power, noise, etc.
In practice, designers are interested in performing \textit{layout} simulations for
measuring the performance metric on any design configuration.
These simulations are however expensive to run; designers instead often turn to \textit{schematic} simulations which return inexpensive proxy metrics correlated with  the target metric. 

In this problem,
the circuit configurations are represented by an 8 dimensional vector, with elements taking continuous values between 0 and 1. 
The offline dataset consists of 1000 pairs of circuit configurations and 3 auxiliary signals including a scalar goodness score based on the 
schematic simulations.
 We consider the same baselines as before. 
We also consider BOX-GP-UCB \cite{perrone2019learning} which confines the search space to a hyper-cube over the promising region based on all auxiliary tasks in the offline data. 
Unlike the considered HPO problems, the offline circuit dataset contains data from more than just one auxiliary task, allowing us to consider BOX-GP-UCB as a viable baseline. MF-GP-UCB was ran with the schematic score as the lower fidelity approximation of the target function.
 We ran each algorithm with $l_\alpha=0.1$ for 100 iterations and measured simple regret against iteration.
As reflected in the regret curves in Figure~\ref{fig:res_ckt}, \name{} outperforms other algorithms. 

\subsection{Ablations}
\label{sec:ablation}
\textbf{Effect of auxiliary tasks.} It is important to analyze how learning on other tasks affects the performance. To this end, we considered the circuit design problem with 1 and 3 auxiliary offline tasks. In Figure \ref{fig:abl_ntask}, 
task 1 (yellow) is the most correlated and task 3 (red) is the least correlated task with the objective function. The regret curves suggest that the performance would be poor if the correlation between tasks is low. Moreover, the features pre-trained on the combination of all three tasks provide more information to the warm-GP than those pre-trained only on one of the tasks.

\textbf{BLR with \name{}'s acquisition function.} A key difference between \name{} and ABLR \cite{perrone2018scalable} is replacing the BLR layer with a GP. 
To show the merits of having a GP,
we ran an experiment on Protein dataset and replaced the GP with a BLR in \name{}'s procedure. Figure~\ref{fig:ablation-blr} shows that \name{} with \warmgp significantly outperforms \name{} with a BLR layer. 

\section{Related Work}
\label{sec:related}
\textbf{Transfer Learning in Bayesian Optimization}:
Utilizing prior information for applying transfer learning to improve Bayesian optimization has been explored in several prior papers. Early work of \cite{swersky2013multi} focuses on the design of multi-task kernels for modeling task correlations~\cite{poloczek2016warm}.
These models tend to suffer from lack of scalability; \cite{wistuba2018scalable,feurer2018scalable} show that this challenge can be partially mitigated by training an ensemble of task-specific GPs that scale linearly with the number of tasks but still suffer from cubic complexity in the number of observations for each task. To address scalability and robust treatment of uncertainty, several prior works have been suggested \cite{ salinas2020quantile, springenberg2016bayesian, perrone2018scalable}.  
\cite{salinas2020quantile} employs a Gaussian Copula to learn a joint prior on hyper-parameters based on offline tasks, and then utilizes a GP on the online task for adapt to the target function.
\cite{springenberg2016bayesian} uses a BNN as surrogates for MBO; however, since training BNNs is computationally intensive \cite{perrone2018scalable} proposes to use a deterministic NN followed by a BLR layer at the output to achieve scalability. 

Some other prior work exploit certain assumptions between the source and target data. For example \cite{shilton2017regret,golovin2017google} assume an ordering of the tasks and use this information to train GPs to model residuals between the target and auxiliary tasks. \cite{feurer2015initializing, wistuba2015learning} assume existence of a similarity measure between prior and target data which may not be easy to define for problems other than hyper-parameter optimization. 
A simpler idea is to use prior data to confine the search space to promising regions \cite{perrone2019learning}. However, this highly relies on whether the confined region includes the optimal solution to the target task. 
Another line of work studies utilizing prior optimization runs to meta-learn acquisition functions \cite{volpp2019meta}. This idea can be utilized in addition to our method and is not a competing direction.

\textbf{Multi-fidelity Black-box Optimization (MFBO)}:
In multi-fidelity scenarios we can query for noisy approximations to the target function relatively cheaply. 
For example, in hyperparameter optimization, we can query for cheap proxies to the performance of a configuration on a smaller subset of the training data~\cite{petrak2000fast}, early stopping~\cite{li2017hyperband}, or by predicting learning curves~\cite{domhan2015speeding,klein2017fast}.
We direct the reader to Section 1.4 in \cite{hutter2019automated} for a comprehensive survey on MBFO. Such methods, similar to MF-GP-UCB~\cite{kandasamy2019multi} (section \ref{sec:exp_ckt}),  
are typically constrained to scenarios where such low fidelities are explicitly available and make strong continuity assumptions between the low fidelities and the target function.


\textbf{Deep Kernel Learning (DKL)}:
Commonly used GP kernels (e.g. RBF, Matern) can only capture simple correlations between points a priori. 
DKL~\cite{huang2015scalable, calandra2016manifold} addresses this issue by learning a latent representation via NN that can be fed to a standard kernel at the output.
\cite{snoek2015scalable} employs linear kernels at the output of a pre-trained NN while \cite{huang2015scalable} extends it to use non-linear kernels.
The \textit{warm-GP} in \name{} can be understood as a DKL surrogate model trained using offline data from auxiliary tasks.


\section{Conclusion}
We proposed \name{}, a no-regret algorithm that employs a careful hybrid of neural networks and Gaussian Processes and a novel acquisition procedure for scalable and sample-efficient Multi-task Bayesian Optimization. 
We derived \name{}'s theoretical regret bound and empirically showed it outperforms other competing approaches on set of real-world optimization problems. 



\small
\bibliography{main}

\newpage
\appendix
\onecolumn
\newpage

\section{Proofs of Theoretical Results}
\label{app:proofs}

We will present proofs for Theorem~\ref{theorem:1} and an additional result in Theorem~\ref{theorem:2} that extends the no-regret guarantees of \name{} to continuous domains.
Our theoretical derivations will build on prior results from \cite{srinivas2010gaussian} and \cite{kandasamy2019multi}.


\subsection{Theorem~\ref{theorem:1}}
Let $\mu^c_{t}(x)$ and $\sigma^c_{t}(x)$ denote the posterior mean and standard deviation of \coldgp{} at the end of round $t$ after observing $\mathcal{D}_{t-1}^{\text{cold}} = \{(x_i, y_i)_{i=1}^{t-1}$\}. Similarly, we will use $\mu^w_{t}(x)$ and $\sigma^w_{t}(x)$ to denote the posterior mean and standard deviation of \warmgp{} at the end of round $t$ after observing $\mathcal{D}_{t-1}^{\text{warm}} = \{(h_\phi(x_i), y_i)_{i=1}^{t-1}\}$.

\begin{lemma}
\label{lemma:1}
Pick $\delta \in (0,1)$ and set $\beta_t = 2 \log\left(\frac{|\chi|\pi_t}{\delta}\right)$ where $\Sigma_{t \ge 1} \pi_t^{-1} = 0.5$, $\pi_t > 0$ (e.g. $\pi_t = \frac{\pi^2 t^2}{3}$). Define $\mu_{t}(x) = \lambda_{t}(x)\mu^c_{t}(x) + (1 - \lambda_{t}(x)) \mu^w_{t}(x)$ and  $\sigma_{t}(x) = \lambda_{t}(x)\sigma^c_{t}(x) + (1 - \lambda_{t}(x)) \sigma^w_{t}(x)$. Then,

\begin{equation*}
    \prob{|f(x) - \mu_{t}(x)| \le \beta_t^{1/2}\sigma_{t}(x), \forall x \in \rchi, \forall t \ge 1 } \ge 1 - \delta.
\end{equation*}

\begin{proof}

Fix $t \ge 1$ and $x \in \rchi$. Based on Assumption~\ref{ass:coldgp}, conditioned on $\mathcal{D}_{t-1}^{\text{cold}}$, $f(x) \sim \mathcal{N}(\mu^c_{t}(x), \sigma^c_{t}(x))$. Similarly, Assumption~\ref{ass:warmgp} implies that conditioned on $\mathcal{D}_{t-1}^{\text{warm}}$, $f(x) \sim \mathcal{N}(\mu^w_{t}(x), \sigma^w_{t}(x))$. Let $\mathcal{A}$ be the event that 
$|f(x) - \mu^c_{t}(x)| \le \beta_t^{1/2}\sigma^c_{t}(x)$ 
and $\mathcal{B}$ the event that 
$|f(x) - \mu^w_{t}(x)| \le \beta_t^{1/2}\sigma^w_{t}(x)$. 
From proof of Lemma 5.1 in \cite{srinivas2010gaussian} we know that given a normal distribution $z \sim \mathcal{N}(0, 1)$, $\prob{z > c} \le 0.5e^{-\frac{c^2}{2}}$. Using $z = \frac{f(x) - \mu^c_{t}(x)}{\sigma^c_{t}(x)}$ and $c=\beta_t^{1/2}$, $\prob{\bar{\mathcal{A}}} \le e^{-\beta_t/2}$. Similarly, $\prob{\bar{\mathcal{B}}} \le e^{-\beta_t/2}$. Using union bound, we have:

\begin{equation*}
    \prob{\bar{\mathcal{A}} \lor \bar{\mathcal{B}}} \le \prob{\bar{\mathcal{A}}} + \prob{\bar{\mathcal{B}}} \le 2e^{-\beta_t/2}.
\end{equation*}

By union bound, we have:
\begin{equation*}
    \prob{\evAb \lor \evBb} \le |\rchi|\sum_{t\ge1} 2e^{-\beta_t/2} \le 
    \delta \ \ \ \ \ \forall x \in \rchi, \forall t \ge 1.
\end{equation*}

The event in this Lemma is just $\mathcal{A} \land \mathcal{B}$ and the proof is concluded. 
\end{proof}
\end{lemma}

Next, we state two lemmas from prior work.

\begin{lemma}
\label{lemma:2}
If $|f(x) - \mu_{t}(x)| \le \beta_t^{1/2}\sigma_{t}(x)$, then $r_t$ is bounded by $2 \beta_t^{1/2}\sigma_{t}(x_t)$.
\begin{proof}
See Lemma 5.2 in \cite{srinivas2010gaussian}. It employs the results of Lemma 2 to prove the statement.
\end{proof}
\end{lemma}



\begin{lemma}
\label{lemma:4}
Let $\sigma^2_t(x)$ denote the posterior variance of a GP after $t-1$ observations, and let $A \subset \rchi$. Assume that we have queried $f$ at $n$ points $(x_t)_{t=1}^{n}$ of which $s$ points are in $A$. Then $\sum_{t:x_t \in A} \sigma_{t}^2(x) \le \frac{2}{\log(1 + \sigma ^ {-2})} \Psi_{s}(A)$.

\begin{proof}
See Lemma 8 in \cite{kandasamy2019multi}.
\end{proof}
\end{lemma}

\paragraph{Proof for Theorem~\ref{theorem:1}}

\begin{proof}
From Lemma~\ref{lemma:2}, we have: 


\begin{align}
    r_t^2 &\le  4\beta_t\sigma_{t}^2(x_t)
\end{align}
Summing over instantaneous regrets for $T$ rounds, we get:
\begin{align}
    \sum_{t=1}^T r_t^2 &\le \sum_{t=1}^T 4\beta_t\sigma_{t}^2(x_t) \\
    \label{eq:inst_reg1}
    &\le 4 \beta_T \sum_{t=1}^T \sigma_{t}^2(x_t) \\
    \label{eq:inst_reg2} 
    &\le 4 \beta_T\left(\sum_{t:x_{t} \in \chi_g} \sigma^{c^2}_{t}(x_t) + \sum_{t:x_{t} \in \bar{\chi}_g} \sigma^{w^2}_{t}(x_t)\right) \\
    \label{eq:inst_reg3} 
    &\le \frac{8\beta_T}{1 + \sigma_n^{-2}} \left(\Psi_{T-s}(\chi_g) +  \Psi_{s}(\bar{\mathcal{Z}}_g)\right)
\end{align}

Eq~\ref{eq:inst_reg1} follows from the monotonicity of 
$\beta_t = 2 \log(\pi_t/\delta)$. 
Eq.~\ref{eq:inst_reg2} follows from the definition of $\sigma_t$ in Lemma~\ref{lemma:1} and the last inequality in Eq.~\ref{eq:inst_reg3} follows from Lemma~\ref{lemma:4}. 

Finally, from Cauchy-Schwartz inequality, we know that $R^2_{T} \le T\sum_{t=1}^Tr^2_t$.
Combining with Eq.~\ref{eq:inst_reg3}, we obtain the result in Theorem~\ref{theorem:1}.
\end{proof}

\subsection{Extension to Continuous Domains}
\label{sec:infinite}

We will now derive regret bounds for the general case where $\rchi \subset [0, r]^d$ is a d-dimensional compact and convex set with $r > 0$. 
This will critically require an additional Lipschitz continuity assumption on $f$.

\begin{theorem}
\label{theorem:2}
Suppose that kernels $\kappa^c$ and $\kappa^w$ are such that the derivatives of \coldgp and \warmgp sample paths are bounded with high probably. Precisely, for some constants $a, b > 0$,

\begin{equation}
    \label{eq:lipschitz}
    P\left\{ \left | \sup_{x\in \rchi} \frac{\partial f}{\partial x_j} \right | > L \right\} \le a e^{-(L/b)^2}, \\ j = 1, 2, \dots, d.
\end{equation}

Pick $\delta \in (0, 1)$, and set $\beta_t = 2 \log(4\pi^2t^2/3\delta) + 4d\log(dtbr \sqrt{\log(4da/\delta)})$. Then, running \name{} for $T$ iterations results in a sequence of candidates $(x_t)_{t=1}^{t=T}$ for which the following holds with probability at least $1 - \delta$:

\begin{equation*}
    R_{T} < \sqrt{CT\beta_T \{\Psi_{T-s}(\rchi_g) + \Psi_{s}(\bar{\mathcal{Z}}_g)\}} + \frac{\pi^2}{6}, \forall T \ge 1
\end{equation*}

where $C = \nicefrac{1}{(1 + \sigma_n^2)}$.
\end{theorem}

To start the proof, 
we first show that we have confidence on all the points visited by the algorithm.

\begin{lemma}
\label{lemma:first_lemma_continuous}
Pick $\delta \in (0, 1)$ and set $\beta_t = 2 \log(\pi_t / \delta)$, where $\sum_{t \ge 1} \pi_t^{-1} = 0.5$, $\pi_t > 0$. Define $\mu_{t}(x) = \lambda_{t}(x)\mu^c_{t}(x) + (1 - \lambda_{t}(x)) \mu^w_{t}(x)$ and  $\sigma_{t}(x) = \lambda_{t}(x)\sigma^c_{t}(x) + (1 - \lambda_{t}(x)) \sigma^w_{t}(x)$. Then, 
\begin{equation*}
    |f(x_t) - \mu_t(x_t)| \le \beta_t^{1/2}\sigma_t(x_t),\ \forall t \ge 1
\end{equation*}

holds with probability of at least  $1 - \delta$.

\begin{proof} 
Fix $t \ge 1$ and $x \in \rchi$. 
Similar to Lemma \ref{lemma:2}, 
$\prob{\bar{\mathcal{A}} \lor \bar{\mathcal{B}}} \le 2 e ^{- \beta_t / 2}$.
Since $e ^{- \beta_t / 2} = \delta / \pi_t$, using the union bound for $t \ge 1$ concludes the statement.
\end{proof}

\end{lemma}

For the purpose of analysis, we define a discretization set $\rchi_t \subset \rchi$, so that the results derived earlier can be re-applied to bound the regret in continuous case. To enable this approach we will use conditions on $L$-Lipschitz continuity to obtain a valid confidence interval on the optimal solution $x^\ast$. Similar to \cite{srinivas2010gaussian}, let us choose discretization $\rchi_t$ of size $\tau_t^d$ (i.e. $\tau_t$ uniformly spaced points per dimension in $\rchi$) such that for all $x \in \rchi$ the closest point to $x$ in $\rchi_t$, $[x]_t$, has a distance less than some threshold. Formally, $||x - [x]_t||_1 \le \nicefrac{rd}{\tau_t}$. 






\begin{lemma}
\label{lemma:regret_continuous}
Pick $\delta \in (0, 1)$ and set $\beta_t = 2 \log(4\pi_t/\delta) + 4d\log(dtbr \sqrt{\log(4da/\delta)})$, where $\sum_{t \ge 1} \pi_t^{-1} = 0.5$, $\pi_t > 0$. Then, for all $t \ge 1$, the regret is bounded as follows:

\begin{equation}
    r_t \le 2 \beta_t ^{1/2}\sigma_t(x_t) + \frac{1}{t^2}
\end{equation}

with probabilty of at least $1 - \delta$.

\begin{proof}
In light of Lemma~\ref{lemma:first_lemma_continuous},the proof follows directly from Lemma 5.8 in \cite{srinivas2010gaussian}. 
\end{proof}
\end{lemma}
\paragraph{Proof of Theorem~\ref{theorem:2}}

\begin{proof}

From Eq.~\ref{eq:inst_reg3} in the proof of Theorem \ref{theorem:1}, we have shown that: 

\begin{equation*}
    \sum_{t=1}^{T} 4 \beta_t\sigma_t^{2}(x_t) \le C\beta_T (\Psi_{T-s}(\rchi_g) + \Psi_s(\bar{\mathcal{Z}}_g))\ \ \ \ \ \forall T \ge 1.
\end{equation*}

Therefore, using Cauchy-Schwarz:

\begin{equation*}
    \sum_{t=1}^{T} 2 \beta_t ^ {1/2} \sigma_t(x_t) \le \sqrt{C\beta_T (\Psi_{T-s}(\rchi_g) + \Psi_s(\bar{\mathcal{Z}}_g))}\ \ \ \ \  \forall T \ge 1.
\end{equation*}

Since $\sum_{t=1}^{T} \frac{1}{t^2} \le \frac{\pi^2}{6}$, Theorem~\ref{theorem:2} follows Lemma~\ref{lemma:regret_continuous}.
\end{proof}

\newpage
\section{Implementation Details}

\subsection{Pre-training Details}

\begin{figure}[h]
    \centering
    \includegraphics[width=0.75\linewidth]{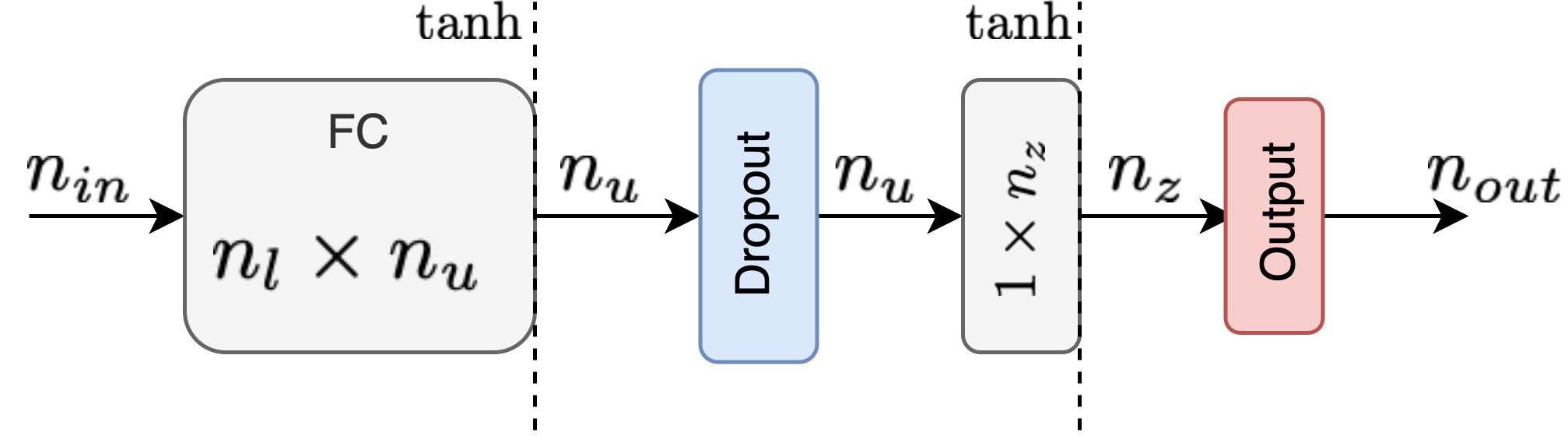}
    \caption{NN Architecture during pre-training: The first blocks is $n_l$ layers of  $n_u$ hidden units with $\tanh$ activations. Following that is a dropout layer and then a single layer perceptron to get $n_z$ features. Thereafter, the latent features are mapped linearly to the output.}
    \label{fig:arch}
\end{figure}

Figure \ref{fig:arch} illustrates the skeleton of the architecture that was used for all experiments. The input configuration is fed to a multi-layer perceptron of $n_l$ layers with $n_u$ hidden units. Then, optionally, a dropout layer is applied to the output and the result is fed to another non-linear layer with $n_z$ outputs. The latent features are then mapped to the output with a linear layer. All activations are $\tanh$. 

For HPO experiments, we have used $n_u=32, n_z=4, n_l=3, \text{learning rate} = 5\times 10^{-5}$, and $\text{batch size} = 128$. For circuit experiments we used $n_u=200, n_z=32, n_l=3, \text{learning rate} = 3\times 10^{-4}, \text{batch size} = 64$, and dropout rate of $0.5$. These hyper-parameters were chosen based on random search by observing the prediction accuracy of the pre-training model on the auxiliary validation dataset which was 20\% of the overall dataset.  

\subsection{Details of training the Gaussian Process hyper-parameters}
For both warm and cold GP, we consider a Matern kernel (i.e. $\kappa(x, x') = \frac{2 ^ {1 - \nu}}{\Gamma(\nu)} (\sqrt{2\nu} r^2) K_\nu (\sqrt{2\nu} r^2)$ with $\nu=2.5$ where $r^2 = \frac{||x - x'||^2}{\theta}$). The length scale $\theta$ and observation noise $\sigma_n$ are optimized in every iteration of BO by taking 100 gradient steps on the likelihood of observations via Adam optimizer with a learning rate of $0.1$. 

\subsection{Acquisition Function Details}
For all experiments, we used Upper Confidence Bound with the exploration-exploitation hyper-parameter at round $t$ set as $10 \exp{\frac{-t}{T}}$ where $T$ is the budget of total number of iterations. This way we favor exploration more initially and gradually drift to more exploitation as we approach the end of the budget. For optimization of acquisition function, we use the derivative free algorithm CMA-ES~\cite{hansen2016cma}.

\subsection{Dealing with Categorical Variables in HPO}
We handle categorical and integer-valued variables in BO similar to \cite{garrido2020dealing}. 
In particular, we used $\kappa^c(T(x), T(x'))$ as the kernel where $T: \rchi \to \mathcal{T}$ is a  deterministic transformation that maps the continuous optimization variable $x$ to a representation space $\mathcal{T}$ that adheres to a meaningful distance measurement. 
For example, for categorical parameters, it converts a continuous input to a one-hot encoding corresponding to a choice for that parameter, and for integer-valued variables, it converts the continuous variable to the closest integer value. Similarly, for the pre-training phase, we also train using $h_\phi(T(x))$.

\newpage
\section{Experimental Evidence}
\label{app:exp}

All the experiments were done on a quad-core desktop.

\subsection{Space compression through the pretrained NN}
In this experiment we studied the latent space of a NN fed with uniformly sampled inputs for circuit design and see that 75\% of data variance is preserved in only 4 dimensions (with $n_z=32$), suggesting that the warm-GP is operating in a compressed space.

\begin{figure*}[h]
    \centering
    \includegraphics[width=0.5\textwidth]{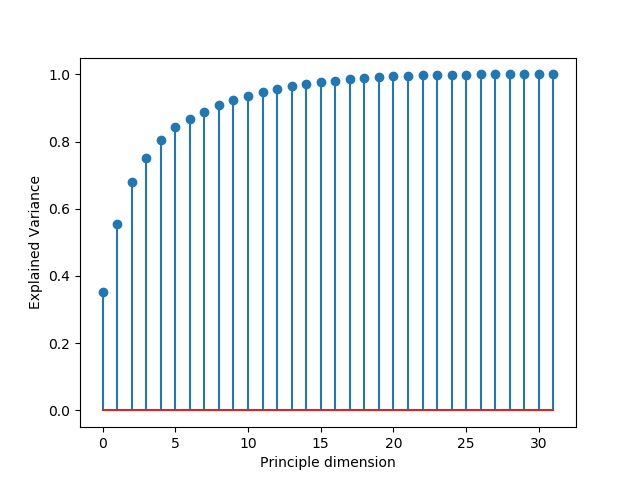}
    \caption{Explained Variance of latent dimensions}
    \label{fig:compression_evidence}
\end{figure*}

\subsection{Discussion: Choice of $l_\alpha$}
The threshold $l_\alpha$ is a key design hyperparameter for defining the acquisition function in \name{}. \name{} with $l_\alpha = \infty$ reduces to GP-UCB (i.e. cold-GP only) and with $l_\alpha = 0$, it reduces to offline DKL (i.e. warm-GP only).

Table \ref{tab:abl_lf} shows the effect of different choices for $l_\alpha$ on the performance of the algorithm for both HPO and circuit design problems. As we can see, small values for $l_\alpha$ (e.g. $\sim 0.01$) cause \name{} to rely more on the accuracy of the warm-GP model and result in sub-optimal convergence in case of model discrepancy between warm-GP and the target task. On the other hand, bigger values of $l_\alpha$ (e.g. $\sim 0.2$) cause \name{} to give more weight to the cold-GP and rely less on prior data. We also note that there is a wide range of $l_\alpha$ that \name{} performs well relative to other baselines, suggesting a good degree of robustness and less tuning in practice.
Even though the optimal choice of $l_\alpha$ really depends on the exact problem setup (e.g. 0.05 for circuits problem, and $0.2$ for Slice localization), we have found that the choice of $l_\alpha=0.1$ is a good initial choice for all the problems considered. 


\begin{table}[h]
\centering
\caption{The average normalized simple regret at the last iteration for different values of $l_\alpha$ (lower is better). The scores are normalized to GP-UCB's simple regret at the last iteration. }
\label{tab:abl_lf}
\resizebox{\textwidth}{!}{%
\begin{tabular}{@{}lllllll@{}}
\toprule
                          & GP-UCB & JUMBO-0.01 & JUMBO-0.05 & JUMBO-0.1   & JUMBO-0.2 & Offline DKL ($l_\alpha=\infty$) \\ \midrule
Protein         & 1.0 $\pm$ 0.08        & 2.09 $\pm$ 0.00         & 1.45 $\pm$ 0.18          & $\mathbf{0.00 \pm 0.00}$ & 0.29 $\pm$ 0.05        & 0.77 $\pm$ 0.13                 \\
Parkinsons & 1.0 $\pm$ 0.05        & 0.52 $\pm$ 0.07         & 0.19 $\pm$ 0.05          & $\mathbf{0.07 \pm 0.05}$ & 0.45 $\pm$ 0.05        & 1.0 $\pm$ 0.02                  \\
Naval          & 1.0 $\pm$ 0.07        & 0.97 $\pm$ 0.07         & $\mathbf{0.46 \pm 0.03}$ & 0.49 $\pm$ 0.04          & 0.78 $\pm$ 0.07        & 1.78 $\pm$ 0.06                 \\
Slice        & 1.0 $\pm$ 0.02        & 5.54 $\pm$ 0.5          & 2.87 $\pm$ 0.12          & 2.69 $\pm$ 0.29          & $\mathbf{0.94 \pm 0.17}$        & 13.77 $\pm$ 0.57                \\
Circuit           & 1.0 $\pm$ 0.06        & 0.16 $\pm$ 0.00          & $\mathbf{0.09 \pm 0.00}$           & 0.18 $\pm$ 0.01           & 0.24 $\pm$ 0.01         & 1.01 $\pm$ 0.01                  \\ \bottomrule
\end{tabular}%
}
\end{table}

\subsection{Ablation: Dynamic choice of $\lambda_t$}
In this ablation we illustrate that the dynamic choice of $\lambda_t$ is indeed better than choosing it to be a constant value. The intuition behind it is that by choosing a constant coefficient we essentially allow the acquisition function to choose points with very high $\alpha^{\text{cold}}$ but low $\alpha^{\text{warm}}$ scores. However, $\alpha^{\text{cold}}$ should not be trusted because of the warm-start problem in BO. 

\begin{figure}[h]
    \centering
    \includegraphics[width=\linewidth]{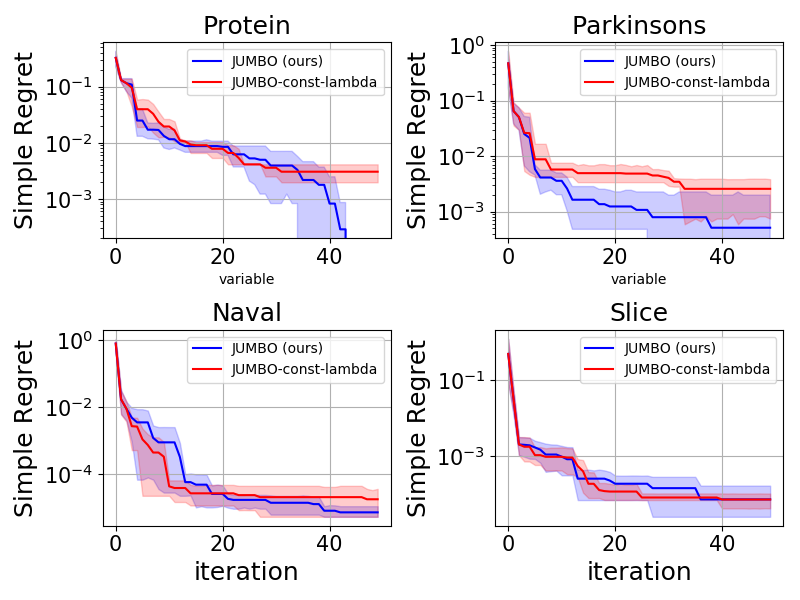}
    \caption{Ablation on the dynamic choice of $\lambda_t$}
    \label{fig:abl_const_lambda}
\end{figure}

Figure \ref{fig:abl_const_lambda} compares \name{} with dynamic and constant $\lambda_t$ on the four HPO problems. It can be seen that \name{} with constant $\lambda_t=0.5$ immaturely reaches a sub-optimal solution in all the experiments. 

\subsection{Tabular Experimental Results}

In this section we present the quantitative comparison between \name{} and the best outstanding prior work for each experimental case. In this table we have also included \textit{BOHB} as another relevant multi-fidelity baseline. 

BOHB combines the successive halving approach introduced in HyperBand~\cite{li2017hyperband} with a probabilistic model that captures the density of good configurations in the input space. Unlike other methods BOHB employs a fixed budget and utilizes the information beyond epoch 3. It runs multiple hyperparameter configurations in parallel and terminates a subset of them after every few epochs based on their current validation error until the budget is exhausted.

\begin{table}[h]
\caption{Comparison of simple regret for HPO. Lower is better. On average \name{}'s simple regret at convergence is 45\% better than the state-of-the-art MBO baseline in each experiment.}
\label{tab:exp_results}
\vspace{1mm}
\centering
\resizebox{0.8\textwidth}{!}{%
\begin{tabular}{@{}rrrrr@{}}
\toprule
             & \multicolumn{1}{l}{Protein ($\times 10^{-3}$)} & \multicolumn{1}{l}{Parkinsons ($\times 10^{-3}$)} & \multicolumn{1}{l}{Naval ($\times 10^{-5}$)} & \multicolumn{1}{l}{Slice ($\times 10 ^ {-4}$)} \\ \midrule
GP-UCB          & 1.98          & 4.93          & 1.4           & 0.77          \\
MT-BOHAMIANN    & 6.71          & 2.13          & 2             & 0.84          \\
MT-ABLR         & 13.52         & 4.91          & 2.3           & 1.42          \\
OfflineDKL      & 1.40          & 2.67          & 4.9           & 10.67         \\
BOHB            & 6.38          & 3.16          & 2.1           & \textbf{0.23} \\
GCP             & 7.50          & 3.15          & 3.3           & 0.46 \\
\name{} (ours)  & \textbf{0}    & \textbf{0.23} & \textbf{0.7}  & 0.73          \\ \bottomrule
\end{tabular}%
}
\end{table}

\end{document}